# A Self-Guided Framework for Radiology Report Generation

Jun Li[1,2], Shibo Li[1(✉)], Ying Hu[1(✉)], Huiren Tao[3]

[1] Shenzhen Institutes of Advanced Technology, Chinese Academy of Sciences
`{jun.li, sb.li, ying.hu}@siat.ac.cn`
[2] University of Chinese Academy of Sciences
[3] Shenzhen University General Hospital

**Abstract.** Automatic radiology report generation is essential to computer-aided diagnosis. Through the success of image captioning, medical report generation has been achievable. However, the lack of annotated disease labels is still the bottleneck of this area. In addition, the image-text data bias problem and complex sentences make it more difficult to generate accurate reports. To address these gaps, we present a self-guided framework (SGF), a suite of unsupervised and supervised deep learning methods to mimic the process of human learning and writing. In detail, our framework obtains the domain knowledge from medical reports without extra disease labels and guides itself to extract fined-grain visual features associated with the text. Moreover, SGF successfully improves the accuracy and length of medical report generation by incorporating a similarity comparison mechanism that imitates the process of human self-improvement through comparative practice. Extensive experiments demonstrate the utility of our SGF in the majority of cases, showing its superior performance over state-of-the-art methods. Our results highlight the capacity of the proposed framework to distinguish fined-grained visual details between words and verify its advantage in generating medical reports.

## 1  Introduction

Medical images are essential to diagnose and find potential diseases. The radiologists' daily task includes analyzing extensive medical images, which supports the doctor in locating the lesions more precisely. Due to the increasing demand for medical images, radiologists are dealing with an unduly high workload. Nevertheless, writing reports is a time-consuming and knowledge-intensive mission. As a result, it's critical to develop an automatic and accurate report generation system to reduce their burden.

Recently, the success of image captioning has become a stepping stone in the development of medical report generation [1-4]. However, radiology images are similar, and the medical reports are diverse, leading to a data bias problem. Additionally, the long sentences with fewer keywords increase the difficulties of predicting reports. To address these problems, some methods [5, 6] used medical terminologies and report subject headings as image labels to distinguish visual features. Yet, radiologists need to relabel the data, failing to lighten their workload.

To fill these gaps, we develop the first, to our knowledge, jointing unsupervised and supervised learning methods to mimic human learning and writing. Our framework includes Knowledge Distiller (KD), Knowledge Matched Visual Extractor (KMVE), and Report Generator (RG). KD simulates human behavior of acquiring knowledge from texts and uses an unsupervised method to obtain potential knowledge from reports. With the knowledge from KD, KMVE learns to extract text-related visual features. This module imitates the process of matching images and texts according to cognition. In the end, RG generates reports according to the image features. This process is similar to human self-improvement through comparative practice. Overall, the main contributions of our method are as follows.

- We present a self-guided framework, a platform of both unsupervised and supervised deep learning that is used to obtain the potential medical knowledge from text reports without extra disease labels, which can assist itself in learning fine-grained visual details associated with the text to alleviate the data bias problem.
- Inspired by human comparative practice, our SGF produces long and accurate reports through a similarity comparison mechanism. This strategy assists the framework in combining global semantics to generate long and complex sentences.
- Several experiments have been conducted to evaluate our framework. The comparison results outperform the state-of-the-art methods, showing the effectiveness of our framework. Furthermore, the results indicate that our approach can learn the visual features of each word and pay more attention to the keywords.

## 2      Related Works

**Image Captioning.** Most of the recent works are based on this classic structure [1], adding attention mechanisms to enhance local visual features [2], or creating a plurality of decoding passes to refine the generated sentences [4]. For example, an adaptive attention method [3] has been proposed to automatically switch the focus between visual signals and language. Lu *et al.* [4] used sentence templates to assist image captioning tasks. Recently, some works [7-9] have been proposed for more specific tasks. An anchor-caption mechanism [9] has been used to generate accurate text, such as numbers and time. Wang *et al.* [8] have developed a new approach to emulate the human ability in controlling caption generation. But all these methods [7-9] require image labels and bounding boxes of objects. It's difficult to transfer these methods in medical report generation because most medical reports do not have disease labels.

**Medical Report Generation.** Researchers have made a lot of attempts and efforts to fill these gaps. By using prior knowledge of chest discovery, Zhang *et al.* [10] established a graph model, and it could be injected into the network to enhance the text decoder. In [11], the graph model and disease keywords were changed to prior and posterior knowledge to assist report generation. Although these efforts have achieved good results, radiologists still need to label the data. When the dataset or disease changes, the data need to be labeled again. Undoubtedly, these methods [10, 11] have not fundamentally solved the data-labeling problem and failed to reduce the burden of radiologists. On the other hand, some strategies [12, 13] were used to enhance the encoder-decoder structure. Li *et al.* [12] applied reinforcement learning to strengthen network training. Wang *et al.* [13] designed a framework with two branches to enhance the fine-grained visual details, whose backbone is CNN-RNN [14]. However, the data-driven RNN structures designed in these works are easily misled by rare and diverse exceptions, and cannot effectively generate abnormal sentences. Recently, the transformer network [15] has laid a solid foundation in neural language processing and computer vision. In [16], a memory-driven unit was designed for the transformer to store the similarity of radiology images. This method only pays attention to the similarity of images while neglecting the diversity of text. In [17], a novel aligned transformer was proposed to alleviate the data bias problem. Yet, this method requires extra disease labels that are not suitable for other diseases.

## 3  Methods

Automatic report generation is a modality-changing process that aims to generate descriptive reports $R = \{y_1, y_2, ..., y_{Nr}\}$ from a set of images $I = \{i_1, i_2, ..., i_{Ni}\}$, where $N_i$ and $N_r$ are the total numbers of reports and images. As shown in Fig. 1, the self-guided framework consists of three parts: Knowledge Distiller (KD), Knowledge Matched Visual Extractor (KMVE), and Report Generator (RG).

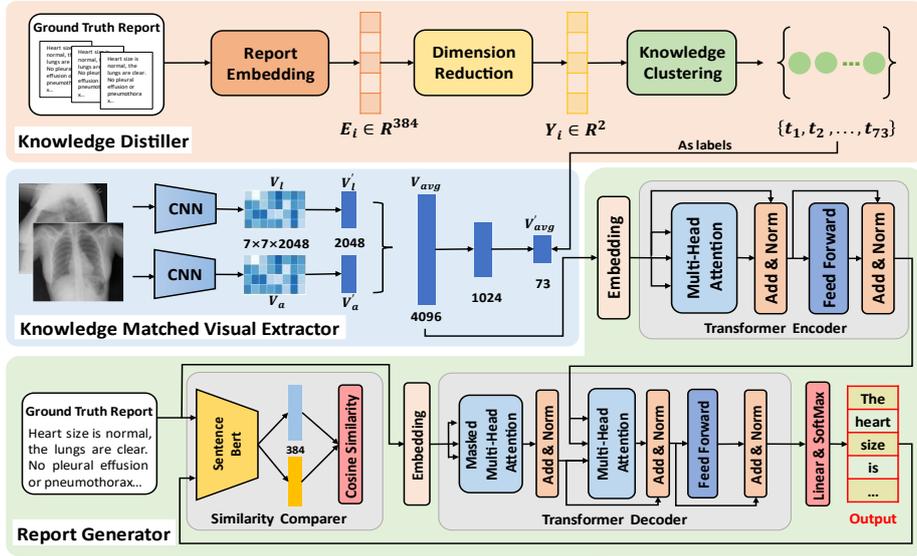

**Fig. 1.** Our proposed self-guided framework includes Knowledge Distiller (KD), Knowledge Matched Visual Extractor (KMVE), and Report Generator (RG).

### 3.1    Knowledge Distiller (KD)

KD is responsible for extracting the potential knowledge from the reports without any extra disease labels. The orange part of Fig. 1 provides an overview of the KD module that contains Report Embedding (RE), Dimension Reduction (DR), and Knowledge Clustering (KC). In RE, the Sentence-Bert (SB) [18] is considered as the embedding method to convert each report to numerical data $E_i \in R^{384}$. The reason for choosing SB as the embedding method in both KD and RG modules is that SB uses Siamese Networks. This structure can derive semantically meaningful sentence embedding vectors which are suitable for clustering and similarity comparison. In addition, SB is a multi-layer bidirectional transformer that has been well pre-trained on two large and widely covered corpora [19, 20], establishing new and most advanced performance in various natural language processing tasks. In DR, the Uniform Manifold Approximation and Projection (UMAP) algorithm [21] is applied for reducing the embedding vector $E_i$, which can be formulated as $Y_i = f_{UMAP}(E_i), Y_i \in R^2$. The number of neighbors and the minimum distance between samples are set to 15 and 0 according to experience, which is robust enough in our training set. Dimension reduction is a common paradigm that can effectively handle the problem of high-dimensional embedding being difficult to cluster and requiring lots of computation in text clustering. Next, the Hierarchical Density-Based Spatial Clustering of Applications with Noise (HDBSCAN) algorithm [22] is used in KC to cluster the set of $Y_i$. HDBSCAN is very suitable for UMAP because UMAP retains many local structures in low-dimensional space, and HDBSCAN will not force clustering these data points. Furthermore, HDBSCAN has fewer and more intuitive hyperparameters and does not require the number of clusters to be specified. In the HDBSCAN algorithm, the minimum cluster size is set to 15, and the metric is set to Euclidean. Finally, the clustering results are marked as $T = \{t_1, t_2, \ldots, t_k\}, k = 73$. To identify whether the clustering result is reliable, we calculate the cosine similarity between each cluster. In Fig. 3 (c), the similarity matrix proves that each topic only shows a high degree of similarity to itself. For each $t_i$, it contains disease information and specific writing styles. The clustering results $T$ can be regarded as the potential knowledge topics from the report.

### 3.2    Knowledge Matched Visual Extractor (KMVE)

KMVE is utilized to alleviate the mismatch feature diversity between the image and text. KMVE takes image pairs $I = \{i_m^1, i_m^2\}, i_m \in R^{N_i}$ as input, where the size of each image is $(3, 244, 244)$. The knowledge topics $T$ from the KD module are adopted as the potential labels of the images. After this operation, the image-text matching process is converted to a classification task. With the knowledge topics as disease labels, the share weights Convolutional Neural Network (CNN) encoders can extract fine-grained visual features associated with text to alleviate the data bias problem. The pre-trained ResNet-101 [23] on ImageNet [24] is used as the backbone of CNN. In the beginning, the CNN extracts anteroposterior and lateral visual features $V_a, V_l \in R^{7 \times 7 \times 2048}$ from the share weights encoder, as shown in Fig. 1. Subsequently, a convolution layer with kernel size $7 \times 7$ and an average pooling layer are added to get the $V'_a \in R^{2048}$ and $V'_l \in R^{2048}$. These two vectors are concatenated together to get the global average features $V_{avg} \in R^{4096}$. To match the size of the knowledge topics $T$, the global average features $V_{avg}$ are reduced to $V'_{avg} \in R^{73}$ by two average pooling layers. In sum, the KMVE loss can be formulated as:

$$\mathcal{L}_{KMVE} = -\sum_{i=1}^{k}\left(t_i \times \log\left(S_f(V'_{avg})\right)\right) \qquad (1)$$

where $t_i$ is the potential label from $T$, $S_f(\cdot)$ is the SoftMax activation function. The global average features $V_{avg}$ retain more information than $V'_{avg}$, and can obtain more fine-grained visual features from images. Therefore, the visual features $V_{avg}$ are chosen as the input of the Report Generator.

### 3.3    Report Generator (RG)

RG is based on Transformer (TF) and Similarity Comparer (SC). TF is consisted with Transformer Encoder (TE) and Transformer Decoder (TD) [15]. In TE, the $V_{avg}$ are converted into query, key, and value at first. Then the Multi-Head Attention (MHA) concatenates all head results together to acquire the visual details in different subspaces. After the MHA, the Residual Connections (RC) [23] and Layer Normalization (LN) [25] are used to fuse features. Besides, the Feed-Forward Network (FFN) is adopted at the end of each block. In TD, the outputs of the TE are converted to key and value. Meanwhile, the embedding matrix of the current input word token is utilized as the query. Like TE, the inputs are changed to vector $h_m$ by the MHA. Next, the $h_m$ is passed to FFN and LN, and it can be described as $h' = L_N\big(h_m + \Phi_{FFN}(h_m)\big)$. Finally, the predicted word is formulated as: $y_t \sim p_t = S_f\big(h'W_p + b_p\big)$, where the $W_p, b_p$ are the learnable parameters. In summary, the TF loss is as follows.

$$\mathcal{L}_{TF} = -\sum_{i=1}^{N_r}(y_i \cdot \log(p_i) + (1-y_i) \cdot \log(1-p_i)) \qquad (2)$$

Due to the long and complex sentences in the report, it is difficult to generate multiple semantic sentences only by the TF module. This is because it only pays attention to the local semantics of the previous words and ignores the overall semantics. Therefore, the SC module is developed to compare the differences between predicted report $p$ and ground truth report $y$, which is helpful for the TF module to learn global semantic information. The SC module is mainly composed of the Sentence Bert, and its settings are the same as the KD module. After embedding, ground truth and predicted report are converted into vectors $y_e \in R^{384}$ and $p_e \in R^{384}$, respectively. Then, the cosine similarities between $y_e$ and $p_e$ are calculated. Besides, the RELU activation is added after the cosine similarity to ensure that the result is above zero. Thus, the similarity score $S$ is simplified as $S = f_{relu}(f_{cs}(y_e, p_e))$, where $f_{relu}$ and $f_{cs}$ are the RELU activation and cosine similarity. In sum, the loss of the SC module is defined as follows.

$$\mathcal{L}_{SC} = -\sum_{i=1}^{N_r} \log(S_i) \qquad (3)$$

## 4      Experiment

**Dataset And Evaluation Metrics.** Our experiments are conducted on the IU-Xray [26] public dataset. IU-Xray is a large-scale dataset of chest X-rays, constructed by 7,470 chest X-ray images and 3,955 radiology reports. The anteroposterior and lateral images associated with one report are used as data pairs. The dataset is split into train-validation-test sets by 7:1:2, and it is the same partition ratio in other methods.

To evaluate the quality of the generated report, the widely used BLEU scores [27], ROUGE-L [28], and METEOR [29] are implemented as the evaluation metrics. All metrics are computed by the standard image captioning evaluation tool [30].

**Implementation Details.** The experiments are conducted on PyTorch with NVIDIA RTX3090 GPU. For MHA in the RG module, the dimension and the number of heads are respectively set to 512 and 8. Our maximum epoch is 50, and the training would stop when the validation loss does not decrease in 10 epochs. The training loss of the entire self-guided framework can be summarized as follows.

$$\mathcal{L}_{SG} = \mathcal{L}_{KMVE} + \mathcal{L}_{TF} + \mathcal{L}_{SC} \qquad (4)$$

we apply the ADAM [31] for optimization and the training rate is set to $5e^{-4}$ and $1e^{-4}$ for the KMVE and RG, respectively. During the training process, the learning rate decays by a factor of 0.8 per epoch, and the batch size is set to 16.

## 5      Results

**Performance Comparison.** In Table 1, our results were compared to other state-of-the-art methods. All metrics reach the top except BLEU-1. The BLEU-1 to BLEU-4 analyze how many continuous sequences of words appear in the predicted reports. In our results, BLEU-2 to BLEU-4 significantly improved. Especially, the BLEU-4 is 4.7% higher than PPKED. One reasonable explanation is that our method ignores some meaningless words and pays more attention to the long phrases used to describe diseases. At the same time, the ROUGE-L is used to investigate the fluency and sufficiency of the predicted report. Our ROUGE-L is 3.9% higher than the previous outstanding method. It means that our approach can generate accurate reports rather than some meaningless word combinations. Furthermore, the METEOR is utilized to evaluate the synonym transformation between predicted reports and the ground truth. The METEOR also shows our framework is effective.

**Ablation study.** Table 2 shows the quantitative results of our proposed methods. It is observed that all evaluation metrics have increased after adding the KMVE module and the SC module separately. The highest metric increased by 5.8% and the lowest metric increased by 1%. Our framework combines the above two tasks to achieve the best performance. It is noteworthy that, the BLEU-1 metric has slightly decreased compared to "BASCE+SC". The reason for this result may be that the SC module is developed for joining the global semantics to overcome the long sentences problem. However, the KMVE module is designed to focus on fine-grained visual features between words. Therefore, the KMVE module might force SGF to pay attention to keywords in sentences, thus reducing the generation of meaningless words.

Table 1. Performance Comparison. Best performances are highlighted in bold.

| Methods | Year | BLEU-1 | BLEU-2 | BLEU-3 | BLEU-4 | ROUGE-L | METEOR |
|---|---|---|---|---|---|---|---|
| Transformer [15] | 2017 | 0.414 | 0.262 | 0.183 | 0.137 | 0.335 | 0.172 |
| H-Agent [12] | 2018 | 0.438 | 0.298 | 0.208 | 0.151 | 0.322 | - |
| CMAS-RL [14] | 2019 | 0.464 | 0.301 | 0.210 | 0.154 | 0.362 | - |
| SentSAT+KG [10] | 2020 | 0.441 | 0.291 | 0.203 | 0.147 | 0.367 | - |
| R2Gen [16] | 2020 | 0.470 | 0.304 | 0.219 | 0.165 | 0.371 | 0.187 |
| PPKED [11] | 2021 | **0.483** | 0.315 | 0.224 | 0.168 | 0.376 | 0.190 |
| SGF | - | 0.467 | **0.334** | **0.261** | **0.215** | **0.415** | **0.201** |

Table 2. Ablation studies. "Base" refers to the vanilla Transformer. "KMVE" and "SC" stand for adding $\mathcal{L}_{KMVE}$ and $\mathcal{L}_{SC}$ loss, respectively. "SGF" refers to using $\mathcal{L}_{SG}$ as the loss. Best performances are highlighted in bold.

| Methods | BLEU-1 | BLEU-2 | BLEU-3 | BLEU-4 | ROUGE-L | METEOR |
|---|---|---|---|---|---|---|
| Base | 0.414 | 0.262 | 0.183 | 0.137 | 0.335 | 0.172 |
| Base + KMVE | 0.462 | 0.306 | 0.221 | 0.165 | 0.353 | 0.182 |
| Base + SC | **0.472** | 0.315 | 0.226 | 0.166 | 0.363 | 0.187 |
| SGF | 0.467 | **0.334** | **0.261** | **0.215** | **0.415** | **0.201** |

**Visualization and Examples.** Fig. 2 shows examples of the generated reports in the test set. The red underlines are the aligned words between ground truth and generated reports. In the first example, neither Transformer nor R2Gen can fully predict complete reports. However, our result is almost consistent with the ground truth. In the second example, the other two methods ignore "non-calcified nodes", and only our prediction shows this lesion. In the third example, the result is still ahead of other methods and hits the most keywords. More importantly, the length of the predicted reports is significantly longer than the other two methods, which proves the proposed framework has distinct advantages in generating complex sentences. And it does not reduce the accuracy in key lesions prediction. Clearly, our method can generate readable and accurate reports from the images, and the results are consistent with Table 1.

**Result Analysis.** In Fig. 3, we analyzed the results from other aspects. Fig. 3 (a) shows the heatmap of each predicted word in the test set. Our model distinguishes each word well and gives different attention to the image. The proposed method successfully attends the fine-grained visual features to describe the disease, such as "clear", "no" and "pneumothorax". For the meaningless words like "the", "are" and ".", our model assigns relatively low probabilities. It proves that KMVE can help the framework to distinguish image details and alleviate the data bias problem. Fig. 3 (b) shows the comparison of cosine similarity and length radio. It can be observed that there is no linear relationship between the report length and similarity score. The increase in the report length will not affect the similarity score. Moreover, the predicted result length is consistent with the ground truth. Besides, the similarity on the test set is basically above 70%. It appears that our framework can overcome the obstacle of generating long and complex sentences in medical reports. Fig. 3 (c) shows the similarity matrix of the KD module in the Knowledge Clustering step. It shows that each topic is highly correlated with itself and proves the reliability of clustering results.

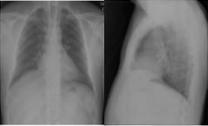

**Fig. 2.** Examples of the generated reports. Our results were compared with the ground truth and the other two methods: Transformer and R2Gen. Highlighted words are consistent parts.

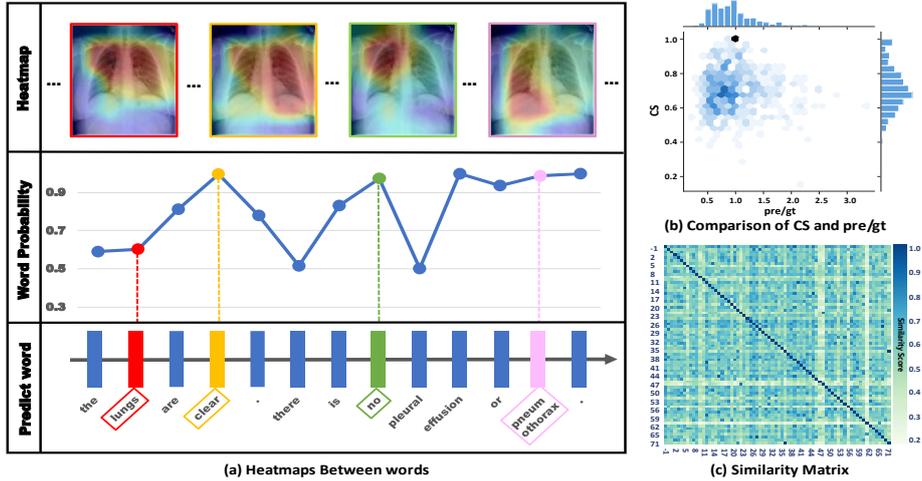

**Fig. 3.** (a) Heatmaps Between words. Each word has different attention and probability in our model. (b) Comparison of cosine similarity and length radio. CS refers to the cosine similarity between predicted reports and ground truths. pre/gt represents the ratio of predicted length to ground truth length. (c) Similarity Matrix. The similarity matrix of the KD module.

## 6     Conclusion

In this work, we propose a self-guided framework to generate medical reports, which imitates the process of human learning and writing. Our approach obtains the domain knowledge in an unsupervised way and guides itself to gain highly text-related visual features to solve the data bias problem effectively. Additionally, the RG can improve the performance through comparative practice, assisting itself in generating complex sentences. Experiment results exceed previous competing methods in most metrics, proving the superior efficiency of our framework. Besides, the result analysis further verifies that our framework can distinguish the fine-grained visual features among words, knowing the differences between keywords and meaningless vocabularies.

**Acknowledgements.** This work was supported in part by Key-Area Research and Development Program of Guangdong Province (No.2020B0909020002), National Natural Science Foundation of China (Grant No. 62003330), Shenzhen Fundamental Research Funds (Grant No. JCYJ20200109114233670, JCYJ20190807170407391), and Guangdong Provincial Key Laboratory of Computer Vision and Virtual Reality Technology, Shenzhen Institutes of Advanced Technology, Chinese Academy of Sciences, Shenzhen, China. This work was also supported by Guangdong-Hong Kong-Macao Joint Laboratory of Human-Machine Intelligence-Synergy Systems, Shenzhen Institute of Advanced Technology.